\documentclass[conference]{IEEEtran}

\IEEEoverridecommandlockouts

\usepackage{times}
\usepackage{epsfig}
\usepackage{graphicx}
\usepackage{amsmath}
\usepackage{amssymb}
\usepackage[nolist]{acronym}
\usepackage{xcolor}
\usepackage{cite}
\usepackage{mathtools}  
\usepackage{tabulary}
\usepackage{lipsum}
\usepackage{bbm} 
\usepackage{comment} 

\usepackage[hidelinks]{hyperref}
\definecolor{darkred}{RGB}{150,0,0}
\definecolor{darkgreen}{RGB}{0,150,0}
\definecolor{darkblue}{RGB}{0,0,150}
\hypersetup{colorlinks=true, linkcolor=red, citecolor=blue, urlcolor=darkblue}

\begin{document}

\title{
Empathy Through Multimodality in Conversational Interfaces
}

\author{  Mahyar Abbasian$^{*1}$, Iman Azimi$^1$, Mohammad Feli$^2$, Amir M. Rahmani$^1$, Ramesh Jain$^1$\\
$^1$University of California, Irvine\\
$^2$University of Turku
%
\thanks{$^*$Corresponding author, {\tt\small abbasiam@uci.edu}}
}

\maketitle
\begin{abstract}
Agents represent one of the most emerging applications of Large Language Models (LLMs) and Generative AI, with their effectiveness hinging on multimodal capabilities to navigate complex user environments. Conversational Health Agents (CHAs), a prime example of this, are redefining healthcare by offering nuanced support that transcends textual analysis to incorporate emotional intelligence. This paper introduces an LLM-based CHA engineered for rich, multimodal dialogue—especially in the realm of mental health support. It adeptly interprets and responds to users' emotional states by analyzing multimodal cues, thus delivering contextually aware and empathetically resonant verbal responses. Our implementation leverages the versatile openCHA framework, and our comprehensive evaluation involves neutral prompts expressed in diverse emotional tones: sadness, anger, and joy. We evaluate the consistency and repeatability of the planning capability of the proposed CHA. Furthermore, human evaluators critique the CHA's empathic delivery, with findings revealing a striking concordance between the CHA’s outputs and evaluators’ assessments. These results affirm the indispensable role of vocal (soon multimodal) emotion recognition in strengthening the empathetic connection built by CHAs, cementing their place at the forefront of interactive, compassionate digital health solutions. 
\end{abstract}

\section{Introduction}

Human conversations transcend mere words, orchestrated as a multimedia experience where tonal inflections, facial dynamics, and gestural semantics are interwoven. These non-verbal cues enrich the emotional and contextual semantics of our exchanges, serving a role analogous to metadata in digital content. Echoing Socrates' ancient apprehensions about written language, we recognize the imperative to resurrect the soul of conversation within our digital interactions.

The advent of mobile technology, replete with sophisticated biometric sensors and capabilities for environmental data capture, has ushered in a transformative shift in communication. Physiological signatures, measured through technologies such as photoplethysmography, accelerometers, and transdermal optical imaging, now provide integral data streams, enriching the field of emotional analytics.

This integration of multimodal sensory data with computational intelligence, especially when interfaced with cutting-edge Generative AI and Large Language Models (LLMs), marks the dawn of a new era in human-computer interaction. Harnessing complex pattern recognition and affective computing capabilities, we envision digital agents capable of providing interactions as nuanced and empathetically resonant as those between humans.

In multimedia computing, the challenge extends beyond crafting algorithms for optimal information fidelity to engineering systems endowed with emotional intelligence. The synergy of LLMs, sensor fusion algorithms, and context-aware computing empowers us to create digital assistants that transcend information delivery to offer genuine relational engagement, fostering trust and personalized user experiences.

Our paper delves into the nexus of empathetic computing and multimedia technology. We investigate the potential of combining LLMs with a suite of contemporary sensing modalities, aiming to develop Conversational Health Agents (CHAs) that redefine traditional paradigms of human-agent interaction. Our goal is to endow these agents with the capacity to decipher and resonate with emotional cues, thus initiating a new chapter in empathetic, human-centric digital communication.

We believe that the multimedia technology community is ready to reconceptualize digital dialogue's future. Our rigorous experimentation and innovation seek to close the gap between technological advancement and authentic human experience, championing AI interactions replete with the depth and empathy synonymous with human connection.

Recent studies have commenced the exploration of LLM-based solutions designed to generate empathetic responses to users' emotional cues. ``CharacterChat'' provided a framework for social support in emotional distress \cite{tu2023characterchat}. Lei et al. \cite{lei2023instructerc} introduced an LLM-based Emotionally Responsive Conversation (ERC) model, utilizing a retrieval template module for contextual relevance. Similarly, Zheng et al. fine-tuned LLaMA for emotional support dialogues \cite{touvron2023llama}, while Nie et al. \cite{nie2024llm} developed a conversational AI therapist incorporating LLMs and smart devices for mental health interventions.

Yet, these early LLM-based approaches are predominantly text-centric, overlooking the vital speech and gestural modalities intrinsic to human interaction. As a result, they fail to capture the full spectrum of contextual and emotional information inherent in conversations, and their responses are limited to textual formats. CHAs, however, stand at the vanguard of LLM and Generative AI applications. Their multimodal capacities are pivotal in navigating intricate user environments, fusing LLMs with diverse external data and AI models to create a holistic experience.

We posit that CHAs have the capability to transcend LLM limitations in conveying empathy by integrating a rich array of multimodal data channels—including textual, speech, video (for facial and gesture analysis), and physiological biomarkers (like heart rate variability). This paper is dedicated to pioneering the integration of speech modalities to surmount these challenges.

In this paper, we introduce an LLM-powered multimodal CHA, designed for rich dialogues within mental health support contexts. This agent discerns emotional cues from speech patterns to provide context-aware and empathetic verbal responses. Utilizing the openCHA framework \cite{abbasian2023conversational}, we integrate an LLM with speech-to-text, speech emotion detection, Internet search, and text-to-speech tools. Our evaluation includes two stages: 1) the consistency and repeatability of the planning capability and 2) inquiring questions with varied emotional tones—sadness, anger, and joy and analyzing responses by human evaluators in terms of empathetic resonance.

\section{Related Work}
In this section, we present an overview of state-of-the-art conversational methods that take emotions into account, including both conventional and LLM-based approaches.

\subsection{Emotion Recognition in Conversation}

There has been a growing interest in leveraging emotion recognition in conversation (ERC) to support mental well-being \cite{raamkumar2022empathetic}. For example, Morris et al. \cite{morris2018towards} proposed a conversational agent aimed at providing empathetic support to users. Their proposed system utilized preexisting emotional support statements drawn from a large corpus of online interactions. In this framework, users shared stressful situations and negative thoughts, receiving feedback selected from the existing corpus of support interactions based on similarity and user ratings.
In another study, Adikari et al. \cite{adikari2022empathic} introduced an conversational agent framework for real-time monitoring and co-facilitation of mental health. This framework consisted of four components: patient emotions analysis using natural language processing (NLP) techniques, group emotion detection employing multiple machine learning approaches, the capture of patient behavioral metrics according to the content shared by the patient within a conversational setting, and a rule-based response generator. Moreover, several studies \cite{poria2019emotion, bertero2016real, ying2019improving} have delved into the utilization of various machine-learning and deep-learning methods to develop ERC models for conversational agents (chatbots).

In addition, studies have harnessed data from multiple modalities to enhance ERC in conversational support systems \cite{fu2023emotion}. Tavabi et al. \cite{tavabi2019multimodal} developed a multimodal deep-learning approach leveraging textual, audio, and visual features to identify opportunities when an agent should convey positive or negative empathetic responses. These modalities were mapped to specific feature representations and then fused for classification. This method achieved an f1-score of 0.71 in discerning when empathetic responses should be delivered by the agent to the user. In a study by Lian et al. \cite{lian2021ctnet}, a multimodal ERC framework was introduced utilizing a transformer-based structure to model intra-modal and cross-modal interactions on word-level lexical and acoustic features. 
The authors further presented two additional multimodal ERC frameworks: a semi-supervised approach utilizing an auto-encoder \cite{lian2022smin}, and an attention-based bi-directional gated recurrent unit (GRU) method \cite{lian2019conversational}.

Despite the advancements, these studies often struggle with open-ended dialogues when the conversation flows freely without specific prompts or questions. These methods are more tailored with predefined responses or prompts, making them better suited for Q\&A interactions in chatbots. Their struggle to handle the complexity and unpredictability of open-ended conversations may limit their effectiveness in providing sufficient mental health support. Additionally, they tend to adopt a discriminative approach, focusing on classification based on predefined categories or labels for emotion recognition. While these methods can identify specific emotions in isolated moments within a conversation, they may overlook emotional tendencies and context that heavily rely on historical utterances. Such approaches prioritize classification accuracy over understanding the emotional expression, potentially leading to misinterpretations or incomplete responses.

 \begin{figure*}[!t]
\includegraphics[width=0.85\linewidth, trim={1.5cm 1cm 11cm 16cm},clip]{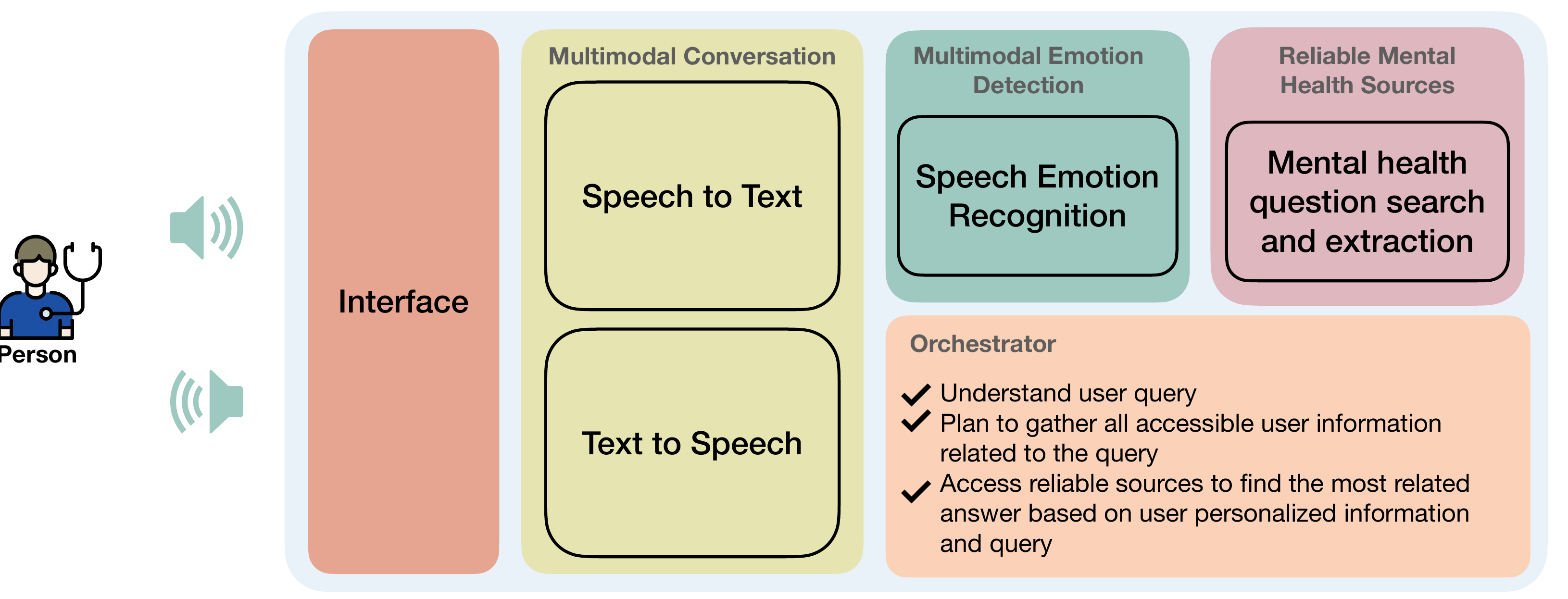}
\centering
\caption {LLM-based CHA for multimodal speech-based emotional support} \label{fig:implementation}
\end{figure*}

\subsection{LLM-powered Conversational Methods}

Recently, LLM-based solutions have been proposed for mental health support, wherein these models aim to generate empathetic responses to users' emotional cues. Tu et al. \cite{tu2023characterchat} proposed CharacterChat, a personalized social support conversation framework for individuals dealing with emotional troubles. Their approach harnessed advanced language models, such as LLaMA \cite{touvron2023llama}, as the response generation backbone, while employing BERT \cite{devlin2018bert} as the memory selection backbone, trained on a dataset created based on MBTI personality types. This study emphasized the importance of interpersonal matching in mental health conversational support systems. In a comparable approach, an LLM-based ERC approach was introduced in \cite{lei2023instructerc}, incorporating a retrieval template module to ensure that the model considers the context of the conversation. Additionally, subject identification and emotion prediction tasks were integrated to model conversation flow and anticipate future emotional tendencies. Zheng et al. \cite{zheng2023building} developed another emotional support conversational system by fine-tuning LLaMA \cite{touvron2023llama} on an emotional support dialogue dataset created using ChatGPT. Furthermore, Nie et al. \cite{nie2024llm} introduced a conversational AI therapist leveraging LLMs and smart devices to address mental health challenges. This platform monitors day-to-day functioning and provides psychotherapeutic interventions through reinforcement learning. 

While these studies \cite{madani2024steering, lissak2024colorful, kang2024can, zhang2024feel, lai2023supporting} have explored LLM-based emotional support conversation for mental health, their objectives were mainly focused on textual data within conversations. This limited focus neglects the potential contributions of other modalities like audio, which can provide valuable insights into users' emotional states and enhance the effectiveness of mental health support systems. Additionally, relying solely on text communication can lead to ambiguity. Textual expressions of emotions might often be ambiguous, making it challenging for models to accurately interpret and respond to users' emotional states. Moreover, textual responses are the primary output of these systems and may not be the most effective means of communication to users in all situations. In contrast to audio or visual modalities, text lacks the richness of tone and other nonverbal cues that can significantly convey empathy and understanding.

\section{LLM-powered Multimodal CHA}

We develop an LLM-powered CHA aimed to offer multimodal emotional support through conversations. This agent leverages LLMs and multimodal conversation and emotion detection modules to interact with users via speech. To achieve this, we adopt an agent-based framework, entitled openCHA \cite{abbasian2023conversational}. Our proposed CHA consists of an orchestrator that interact with these components to generate empathetic responses based on the user's emotion.
For example, when users inquire about mental health issues, the CHA assesses their emotional state by analyzing vocal cues and tailors responses accordingly. These responses are then delivered in speech format. Our proposed CHA comprises five key components: Interface, Multimodal Conversation, Orchestrator, Multimodal Emotion
Detection, and Reliable Mental
Health Sources (see Figure \ref{fig:implementation}). We outline these components in the following.

\subsection{Interface}

The interface facilitates multimodal interactions through a web chat, offering features for voice recording and playback. Recorded voice messages are sent to the Multimodal Conversation component, where they are transcribed into text for further analysis. Additionally, the final response is conveyed back to the interface via the Multimodal Conversation component, where it is converted into spoken voice.

\subsection{Multimodal Conversation}

Multimodal Conversation component plays a crucial role in facilitating multimodal communication within the CHA. It consists of two primary modules, as speech-to-text and text-to-speech, to facilitate speech-enabled conversations. 

For speech-to-text, our tasks utilize the openAI's whisper-base model \cite{whisperwebsite}. Whisper is a versatile speech recognition model that stands out for its ability to handle a range of speech processing tasks, including multilingual recognition, translation, and language identification, thanks to its training on a vast and varied dataset. It employs a Transformer sequence-to-sequence framework, where tasks like speech recognition across multiple languages, translation, language identification, and voice activity detection are integrated into a single workflow.

Meanwhile, for text-to-speech, we leverage the gTTS GitHub library \cite{gtts}. gTTS, is a versatile Python library and command-line tool that connects to Google Translate's text-to-speech API. It features a customizable sentence tokenizer tailored for speech, enabling it to read texts of any length while maintaining correct intonation, handling abbreviations, decimals, and other nuances. Additionally, it offers customizable text pre-processors that can adjust pronunciation as needed.

\subsection{Orchestrator}

The Orchestrator sits at the heart of our CHA, tasked with problem-solving, devising action plans, and generating responses tailored to the user based on their inquiries. This element collaborates with the Multimodal Emotion Detection component to capture the most relevant user information related to the query dynamically. It also collects data from the Reliable Mental Health Sources component to secure the latest and most personalized information. Moreover, by synthesizing the collected data and utilizing LLMs, it extracts insights to formulate the user's final response in text format. The Orchestrator have four important capabilities as Planning, Execution, Short-term Memory, and Response Generator.

The \textit{Planning} capability, fueled by an LLM, acts as the decision-making and cognitive nucleus of the Orchestrator. It is responsible to collate all the information needed to resolve user inquiries effectively. This involves analyzing the user's question to identify the necessary steps for execution we call them tasks.

\begin{figure*}[!t]
\includegraphics[width=\linewidth, trim={1cm 1cm 13cm 0.5cm},clip]{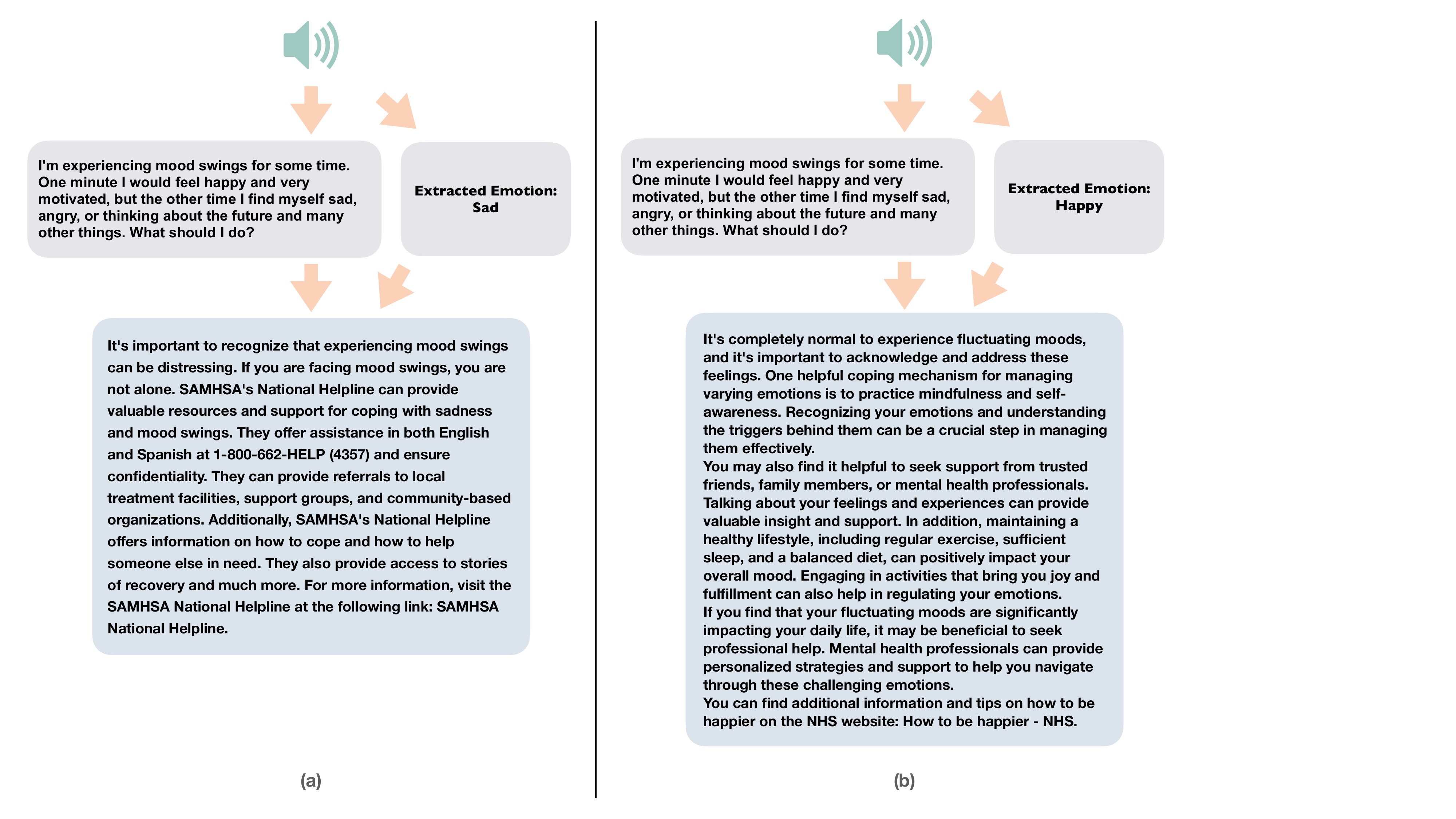}
\centering
\caption {Examples of developed CHA answering a user voice query with Sad (a) and Happy (b) emotions} \label{fig:sample}
\end{figure*}

To convert user questions into actionable tasks, we employ the Tree of Thought \cite{yao2023tree} prompting techniques for strategic planning. This method requires the LLM to undertake three key actions: first, to devise three separate strategies, each a sequence of tasks with specified inputs; second, to outline the pros and cons of these strategies; and third, to determine the most suitable strategy for the query at hand. For the implementation, we use OpenAI's \cite{chatgptwebsite} GPT-3.5-turbo model.

The \textit{Execution} phase within the Orchestrator implements the tasks outlined during the planning stage. These tasks encompass determining the user's current emotional state and conducting searches for information that align with the user's query and perceived emotional condition. The sequence in which these tasks are carried out, along with the specifics of what information to seek, are directed by the Planning capability of the Orchestrator.

The \textit{Short-term Memory} acts as a storage for information gathered from the \textit{Multimodal Emotion Detection} and \textit{Reliable Mental Health Sources} components throughout conversational interactions, essential for enabling multimodality. It holds onto intermediate data that might be too voluminous or complex for the \textit{Planning} LLM or the \textit{Response Generator's} LLM to process directly.

The \textit{Response Generator}, powered by an LLM, utilizes information compiled by the \textit{Planning} component to craft clear and empathetic responses personalized to the user. For this module, we employ OpenAI's GPT-3.5-turbo model, serving as the foundational LLM for the \textit{Response Generator}. 

For the Orchestrator implementation, we customize and leverage the architecture presented in \cite{openchagit}.

\subsection{Multimodal Emotion Detection}

This component enables multimodality within LLM-healthcare integration, enhancing trustworthiness, personalization, and empathy through data insight extraction \cite{topol2023artificial, han2023randomized}. Given the limitations of LLMs in extensive computations, this component will facilitate emotion extraction from various sources like video, audio, or biomarkers analysis. Our current implementation contains only speech emotion recognition.

Our approach for identifying emotional states in speech utilizes the wav2vec2 \cite{baevski2020wav2vec} model, fine-tuned on the IEMOCAP dataset \cite{busso2008iemocap}. This model, known for learning from speech audio to outperform existing semi-supervised methods with greater simplicity, is applied to recognize emotions from speech. The IEMOCAP dataset \cite{busso2008iemocap}, enriched with facial expressions and hand movements data from actors in various emotional scenarios, supports this fine-tuning. We adopt the SpeechBrain \cite{speechbrain} version of wav2vec2 \cite{baevski2020wav2vec}, specifically adjusted with the IEMOCAP \cite{busso2008iemocap} dataset, to serve as our speech emotion detection tool.

\subsection{Reliable Mental Health Sources}

This component retrieves the latest and most relevant data from healthcare sources like healthcare literature and reputable websites through search engines \cite{nori2023can, hiemstra2009information, turtle1992comparison, norican} to avoid hallucination and bias. We incorporated Google Search API (called SerpAPI \cite{serpapi}) and Playwright \cite{playwright} Extract Text to conduct website searches based on user queries and extract relevant context to accurately address users' questions. For more details see \cite{abbasian2023conversational}.


\section{Demonstration and Evaluation}

In this section, we evaluate the performance of our proposed CHA in providing empathetic responses to user queries based on the perceived emotion from their voice. Our aim is to assess the CHA's capability to tailor responses according to the user's current emotional state.

Figure \ref{fig:sample} indicates two examples of how user interaction unfolds with our system, detailing the process from initial voice query to the final audio response tailored to the user's emotional state. Initially, the user's voice query is captured and converted into text by the Speech To Text component. This text is then forwarded to the Orchestrator, which coordinates the planning and formulation of an appropriate response. The planning involves using the Speech Emotion Recognition component to detect emotion from the speech. Once the emotion is identified, it is used to retrieve relevant and reliable answers from the Internet sources. These answers are then sent to the Response Generator, which crafts the final response that is subsequently converted back into audio for the user. Figure \ref{fig:sample}.a illustrates the response generated when the user's emotion is identified as ``Sad,'' directing them towards serious support resources to help mitigate any potential harm. Figure \ref{fig:sample}.b, on the other hand, shows a different response suited to a ``Happy'' emotional state, where the CHA uses a motivational tone and suggests resources to help the user address the issue.

For the evaluation, we chose five neutral questions related to mental health (see Table \ref{tab:questions}). The questions were also tagged as neutral in tone by GPT-3.5. Our evaluation of CHA includes two steps.

In the first step, we focused on the consistency and repeatability of our CHA planning capability in task selection. This involved the planner's ability to accurately extract user emotion from voice data and to search and retrieve information relevant to the user's query. To do this, we input a randomly selected question, chosen from the five voice questions infused with one of three emotions, into the CHA. We repeated this process 500 times.

The performance of the planner is measured using two metrics. The first metric examined how often, out of the 500 tests, the planner successfully identified the emotional state from the voice and retrieved related information pertinent to the user's query. The second metric focused more narrowly on the planner's use of the extracted emotion to conduct Internet searches for relevant data. We obtained the accuracy of \textbf{\%89} and \textbf{\%61} for the first and second metrics, respectively.

\begin{table}[!t]
    \centering
\caption{List of five questions used for the evaluation}
\label{tab:questions}
\resizebox{\linewidth}{!}{
    \begin{tabular}{|c|l|} \hline 
         & I've noticed that I've been experiencing \\
         Question 1&some difficulty concentrating lately. \\
         &Could this just be due to stress, or should\\
         & I be concerned about something more?\\ 
         \hline 
         & I've been feeling a bit more irritable than\\
         Question 2& usual lately, especially at work. Could this\\ 
         &be a sign of burnout, or is it just a phase?\\         
         \hline 
         & I've been experiencing some difficulty \\
         &sleeping, but I'm not sure if it's related to\\
         Question 3&stress or if there could be other underlying\\ 
         &causes. How can I determine the root cause?\\
         \hline 
         & I've been feeling a bit disconnected from my\\
         Question 4& emotions lately. Are there any exercises or \\
         &practices I can try to become more in tune\\
         & with how I'm feeling?\\ \hline 
         & I've been feeling overwhelmed by the \\
         &constant stream of negative news lately.\\
        Question 5& How can I maintain a healthy balance\\
        &between staying informed and protecting\\
        & my mental well-being?\\
         \hline
\end{tabular}
}   
\end{table}

Note that we observed that in cases where the extracted emotion was not used to guide Internet searches, the planner still forwarded the emotional data along with the search results to the response generator. The response generator then used this information to craft an empathetic response. However, when the planner did engage in more sophisticated, emotion-informed searching, it performed a more targeted search based on both the user's query and their emotional state to fetch more personalized information. Regardless of the approach, the final response was tailored to reflect the user's emotional state. Any deviations from these two defined planning paths were considered incorrect. 




\begin{table}[!t]
    \centering
    \caption{Scores from human evaluators that reflect how well the responses align with and show empathy towards the user's questions and Happy, Sad, and Angry emotional states}
    \label{tab:results}
    \begin{tabular}{|c|c|c|c|} \hline 
         &  Happy&  Sad& Angry\\ \hline 
         Question 1&  6&  8.3& 6\\ \hline 
         Question 2&  6.3&  6.3& 7.6\\ \hline 
         Question 3&  5.3&  6& 5.3\\ \hline 
         Question 4&  5.6&  8.6& 6.6\\ \hline 
         Question 5&  8&  7& 7.3\\ \hline 
         Total Average&  6.24&  7.24& 6.56\\ \hline
    \end{tabular}
    
\end{table}

In the second evaluation step, our goal is to measure how well the responses matched the emotional state of the user and the level of empathy they conveyed, given the identified question and extracted user emotion. We asked the five questions to our CHA with three different emotional tones: Happy, Sad, and Angry. This was done to assess how the emotional tone of a question influences CHA's responses.

Then, three external evaluators have reviewed each response, scoring each on how well it aligned with the user's current emotional state and its empathetic quality on a scale from 0 to 10. A score of 0 meant there was no alignment or empathy, and a score of 10 indicated a high degree of both. We then calculated the average scores for each answer-emotion pair. These averages, which reveal the performance of responses across each emotional category, are detailed in Table \ref{tab:results}. The evaluators agreed that scores of 6 or higher indicated a reasonable alignment. Additionally, they reported that responses to questions posed in a state of sadness were more empathetic and aligned compared to those in the other two emotional states.

Consequently, the effective planning capability and commendable evaluation scores for responses indicate the success of the proposed CHA in delivering empathetic answers based on user's emotional state. Our future work will extend the capabilities of CHAs to embrace a broader spectrum of computational empathy. This research has laid the foundation by integrating the first two major modalities of user interactions (i.e., text and voice) providing nuanced emotional understanding and response. We will include other modalities to capture facial and physiological cues and integrate them into CHA's responses. This integration will enable a more holistic empathetic communication framework, driving us towards the objective of CHAs that can engage with and support users with a depth and sensitivity akin to human caregivers. 

\section{Conclusion}

Our exploration into the realm of multimodal CHAs using LLMs offers a promising avenue towards revolutionizing human-computer interaction. 
In this paper, we introduced an LLM-powered multimodal CHA, tailored for in-depth dialogues within health support environments. This agent was capable of interpreting emotional cues from speech patterns to provide context-aware and empathetic verbal responses. Employing the openCHA framework, we integrated an LLM with speech-to-text, speech emotion detection, Internet search, and text-to-speech tools. Our evaluation was conducted in two stages. We, first, assessed the planning capabilities of the agent. Our findings showed that the planner obtained \%89 accuracy to identify the emotional state from the voice and retrieve related information pertinent to the user's query. It also obtained \%61 accuracy to correctly call the Internet searches tool based on the emotion states. Then, we evaluated the responses in terms of empathy. We posed questions with varied emotional tones (i.e., sadness, anger, and joy) and analyzed the responses with the assistance of external human evaluators for empathetic resonance. The external evaluators confirmed that the empathy of the response had reasonable alignment with the three emotions. We observed that responses to questions asked in a state of sadness were deemed more empathetic and better aligned with user expectations compared to those in other emotional states.

\bibliographystyle{IEEEtran}
\bibliography{sample-base}

\end{document}